\documentclass[11pt,a4paper]{article} 

\usepackage{latexsym}
\usepackage{amssymb} 
\usepackage{amsmath} 
\usepackage{amsthm}
\usepackage{eucal} 
\usepackage{bezier}
\usepackage{graphics}
\usepackage[matrix,curve,arrow,tips,frame]{xy}
\usepackage{color}

\theoremstyle{plain} \newtheorem{theorem}{Theorem}

\newtheorem{corollary}[theorem]{Corollary} \newtheorem{cl}{Claim}
\newtheorem{conj}{Conjecture} 

\theoremstyle{definition} \newtheorem{definition}[theorem]{Definition}

\theoremstyle{remark}

\DeclareMathOperator{\PP}{\mathcal{P}}

\newcommand{\X}{\mathcal{X}}
\newcommand{\Y}{\mathcal{Y}}

\DeclareMathOperator{\Since}{\mathsf{Since}}

\DeclareMathOperator{\PREV}{\unitlength 1.00pt\begin{picture}(8.00,10.00)\put(4.00,3.00){\circle*{6}}\end{picture}}
\DeclareMathOperator{\PDIA}{\blacklozenge}
\DeclareMathOperator{\PBOX}{\blacksquare}

\newcommand{\TL}{\mathrm{TL}}
\newcommand{\M}{\mathcal{M}} 

\newcommand{\conn}[9]{
\begin{array}{|c|ccc|} \hline
\multicolumn{4}{|c|}{x #1 y}\\ \hline\hline x\diagdown y &0 &1 &\bot\\\hline
 0 &#2 &#3 &#4\\ 1 &#5 &#6 &#7\\ \bot &#8 &#9
&\bot\\ \hline
\end{array}}

\newcommand{\SAC}{\mathrm{SAC}} 
\newcommand{\GNW}{\mathrm{GNW}}

\newcommand{\Sch}{\mathrm{Sch}}

\newcommand{\lra}{\leftrightarrow}

\newcommand{\true}{1} 
\newcommand{\false}{0}

\newcommand{\A}{\mathfrak{A}}
\newcommand{\B}{\mathfrak{B}}
\newcommand{\E}{\mathcal{E}}

\newcommand{\fM}{\mathfrak{M}}

\newcommand{\C}{\complement}
\newcommand{\trzy}{{\text{\boldmath$\mathsf{3}$}}}
\newcommand{\dwa}{{\text{\boldmath$\mathsf{2}$}}}

\newcommand{\PC}{\mathcal{PC}}
\newcommand{\CC}{\mathcal{C}}

\newcommand{\ce}[1]{[\![#1]\!]}

\newcommand{\TT}{{(\TL|\TL)}}
\newcommand{\pr}{\Pr\nolimits}

\newcommand{\TRUE}{\mathit{true}}
\newcommand{\FALSE}{\mathit{false}}
\newcommand{\atone}{\mathbf{@}_1}
\newcommand{\attwo}{\mathbf{@}_2}

\title{The Temporal Calculus\\ of Conditional Objects and Conditional
Events}

\author{
Jerzy Tyszkiewicz$^{1,2}$\\
Arthur Ramer$^2$\\
Achim Hoffmann$^2$}
\CompileMatrices
\begin{document} 
\begin{titlepage}
\maketitle 
\begin{center}$^1$ Institute of Informatics,\\ University of
Warsaw,\\ Banacha 2,\\ 02-097 Warszawa,\\ Poland.\\ E-mail
{\tt jty@mimuw.edu.pl}.\\ Supported by the Polish Research Council KBN
grant 8 T11C 027 16.\\[5pt]  $^2$ School CSE,\\ UNSW,\\ 2052 Sydney,\\
Australia.\\ E-mail {\tt \{jty|ramer|achim\}@cse.unsw.edu.au}.\\ Supported by
the Australian Research Council ARC grant A 49800112 (1998--2000).
\end{center}
\thispagestyle{empty}
\end{titlepage}
\begin{titlepage}
\vspace{5cm}
\begin{abstract} 
    We consider the problem of defining conditional objects $(a|b),$
which would allow one to regard the conditional probability $\Pr(a|b)$
as a probability of a well-defined event rather than as a shorthand
for $\Pr(ab)/\Pr(b).$ The next issue is to define boolean combinations
of conditional objects, and possibly also the operator of further
conditioning.  These questions have been investigated at least since
the times of George Boole, leading to a number of formalisms proposed
for conditional objects, mostly of syntactical, proof-theoretic vein.

    We propose a unifying, semantical approach, in which conditional
events are (projections of) Markov chains, definable in the
three-valued extension $\TT$ of the past tense fragment of
propositional linear time logic ($\TL$), or, equivalently, by
three-valued counter-free Moore machines. Thus our conditional objects
are indeed stochastic processes, one of the central notions of modern
probability theory. 

Our model precisely fulfills early ideas of de Finetti \cite{d72},
and, moreover, as we show in a separate paper \cite{cea}, all the
previously proposed algebras of conditional events can be
isomorphically embedded in our model.
\end{abstract}
\end{titlepage}

\begin{titlepage}
\tableofcontents 
\end{titlepage}

\section{Preliminaries and statement of the problem}

\subsection{The problem of conditional objects}

Probabilistic reasoning \cite{p88} is the basis of Bayesian methods of
expert system inferences, of knowledge discovery in databases, and in
several other domains of computer, information, and decision
sciences.  The model of conditioning and conditional objects we discuss
serves equally to reason about probabilities over a finite domain $X$,
or probabilistic propositional logic with a finite set of atomic
formulae.

Computing of conditional probabilities of the form
$\Pr(X|Y_1,\dots,Y_n)$ and, by extension of conditional beliefs, is
well understood.  Attempts of defining first the {\em conditional
objects\/} of the basic form $X|Y$, and then defining $\Pr(X|Y)$ as
$\Pr((X|Y))$ were proposed, without much success, by some of the
founders of probability \cite{b57,d72}.  They were taken up
systematically only about 1980.  The development was slow, both
because of logical difficulties \cite{l76,cccp1,cccp2}, and even more
because the computational model is difficult to construct.  (While
$a|b$ appears to stand for a sentence `if $b$ then $a$', there is no
obvious calculation for $\Pr(a|(b|c))$, nor intuitive meaning for
$a|(b|c),$ $(a|b)\land(c|d),$ and the like.)

The idea of defining conditional objects was entertained by some
founders of modern probability \cite{b57,d72}, but generally abandoned
since introduction of the measure-theoretic model.  It was revived
mostly by philosophers in 1970's \cite{a86,v77} with a view towards
artificial intelligence reasoning.  Formal computational models came
in the late 1980's and early 1990's \cite{c87,gnw91,g94}. Only a few
of them have been used for few actual calculations of conditionals and
their probabilities whose values are open to questions \cite{c94,g94}.

In this paper we want to give a rigorous (and yet quite natural and
intuitive) probabilistic and semantical construction of conditionals,
based on ideas proposed by de Finetti over a quarter a century ago
\cite{d72}. It appears that this single formalism contains fragments
precisely corresponding to all the previously considered algebras of
conditional events \cite{cea}.  Seen as a whole, it can be therefore
considered as their common generalisation and perhaps {\em the\/}
calculus of conditionals. 

Our system consists of three layers: the logical part is a three
valued extension of the past tense fragment of propositional linear
time logic, the computation model are three-valued Moore machines (an
extension of deterministic finite automata), and the probabilistic
semantics is provided by three-valued stochastic processes, which
appear to be projections of Markov chains.

\subsection{The main idea}

\paragraph{The main idea.} The main idea of our approach can be seen
as an attempt to provide a precise mathematical implementation of the
following idea of de Finetti \cite[Sect.\ 5.12]{d72}:

\begin{quote}
\begin{sl}
``In the asymptotic approach, the definition of conditional
probability appears quite naturally; it suffices to repeat the
definition of probability \textup{(}as the limiting
frequency\textup{)}, taking into consideration only the trials in
which the conditioning event \textup{(}hypothesis\textup{)} is
satisfied. Thus, $P(E|H)$ is simply the limit of the ratio between the
frequency of $EH$ and the frequency of $H.$ If the limiting frequency
of $H$ exists and is different from zero, the definition is
mathematically equivalent to the compound probability theorem
$P(E|H)=P(EH)/P(H).$ But even if the frequency of $H$ does not tend to
a limit, or the limit is zero, $P(E|H)$ can nonetheless exist
\textup{(}trivial example: $P(H|H)$ is always equal to \/
$1$\textup{)}.''
\end{sl}
\end{quote}

We believe that our attempt is successful: our system will have all
the properties predicted by de Finetti, and, moreover, as we show in a
separate paper \cite{cea}, subsumes all the previously existing
formalisms developed to deal with conditionals, and, finally, appears
to be able to handle some well-known paradoxes of probability in an
intuitive and yet precise manner.

\paragraph{Three truth values.} 

To be able to take into account only the trials in which the
hypothesis is satisfied, one has to introduce a third logical
value. Informally, if one considers two players\footnote{This sounds
definitely better than gamblers ;-).}: one betting $(a|b)$ will hold,
and the other it will not, if in a random experiment (dice toss, coin
flip) $b$ doesn't hold, the game is drawn.  The previous works
considered it to be an evidence that the definition of conditionals
must be necessarily based on many valued logics, the typical choices
being three valued.

Note however, that assigning probability to a three-valued $c$ is
something like squeezing it to become two-valued. For one then assumes
it to be true $\Pr(c)$ of time and false $1-\Pr(c)$ of time, and the
time when $c$ has the third value, typically described as {\em
undefined}, is lost. So, unlike most of our predecessors, we attempt
to preserve the three-valuedness of conditionals as a principle, and
define their probability only on the top of that.

\paragraph{Bet repetitions.}
Now, we should allow the players to repeat their bets. Here, unlike
most of the previous works, if the players repeat the game, we allow
them to bet on properties of the {\em whole\/} sequence of outcomes,
not just the last one.

This is not uncommon in many random experiments, that the history of
the bets influences the present bet somehow.

We present three natural examples, which are natural and have a simple
description.  

The first possibility is that after each bet we we start over---after
the result of the experiment is settled, the (temporal) history is
started anew, the next experiment not taking the old results into
account.

The second is just the opposite---always the entire history, including
earlier experiments, is taken into account.

The third is that no repetition is allowed: after the first experiment
is settled, its outcome is deemed to persist forever, and future
trials are effectively null.  (Regardless of each subsequent element
drawn the result is always defined and remains the same.)

Roughly speaking, the first choice is adopted in bridge, the second in
blackjack and the third in Russian roulette.

This suggests that a conditional isn't merely an experiment with three
possible outcomes. It is indeed a {\em sequence} of experiments, and
the third logical value, often described as {\em unknown}, is often
{\em not y\underline{et} known}. It is clearly a temporal concept, and
thus we are going to consider conditionals as {\em temporal objects.}
This temporal aspect is clearly of {\em past tense\/} type --- the
result of a bet must depend on the history (including present) of the
sequence of outcomes, only.

It is worth noting that there are other approaches which consider
implicitly bet repetition in the modelling of conditionals. These
include \cite{v77,mcgee,g94,ramer}.

\paragraph{Summary.} 
What we undertake is thus the development of a calculus of conditional
objects identified with temporal rules, which, given a sequence of random
elements from the underlying domain, decide after each of the drawn
elements if the the conditional becomes defined, and if so, whether it
is true or false.

We stipulate that, for any reasonable calculus of conditionals, {\em
forming boolean combinations of conditionals, as well as iterated
conditionals, amounts to manipulating on these rules.}

This claim is indeed well motivated: if we fail to associate such rule
to a complex conditional object, we do not have any means to say, in a
real-life situations, who wins the bet on this conditional and when.
So to say, such a conditional would be nonprobabilistic, because one
couldn't bet on it!

\paragraph{Novelty of our approach.}  We would like to stress that
virtually none of the results we prove below is entirely new. Most of
them are simple extensions or reformulations of already known
theorems, as the reader can verify in Section \ref{related}. The
novelty of our approach lies almost entirely in the way we assemble
the results to create mathematically precise representation of an
otherwise quite clear and intuitive notion. And indeed, we feel very
reassured by the fact that we didn't have to invent any new
mathematics for our construction.  Similarly the proofs we give in
this paper are quite straightforward. This is exactly the emergence of
previously-unheard-of complicated algebraic structures (dubbed {\em
conditional event algebras\/} in \cite{kniga}), which prompted us to
have a closer look at conditional events and search for simpler and
more intuitive formalisations. Note that probabilists and logicians
have been doing quite well without conditional events for decades,
which strongly suggests they have had all the tools necessary to use
conditionals in an implicit way for a long time already. To the
contrary, in the emerging applied areas, and in particular in AI,
there is a strong need to have conditional events explicitly present,
and this is why we believe in the importance of our results.

\section{The tools}
\subsection{Pre-conditionals}

Let $\E=\{a,b,c,d,\dots\}$ be a finite set of basic events, and let
$\Sigma$ be the free Boolean algebra generated by $\E,$ and $\Omega$
the set of atoms of $\Sigma.$ Consequently, $\Sigma$ is isomorphic to
the powerset of $\Omega,$ and $\Omega$ itself is isomorphic to the
powerset of $\E.$ Any element of $\Sigma$ will be considered as an
event, and, in particular, $\E\subseteq\Sigma.$

The union, intersection and complementation in $\Sigma$ are denoted by $a\cup
b,\ a\cap b$ and $a^\C,$ respectively.  The least and greatest elements of
$\Sigma$ are denoted $\varnothing$ and $\Omega,$ respectively. However,
sometimes we use a more compact notation, replacing $\cap$ by juxtaposition.
When we turn to logic, it is customary to use yet another notation: $a\lor b,\
a\land b$ and $\lnot a,$ respectively. In this situation $\Omega$ appears as
$\TRUE$ and $\varnothing$ as $\FALSE,$ but $1$ and $0,$ respectively, are
incidentally used, as well. Generally we are quite anarchistic in our
notation, as long as it does not create ambiguities.

We introduce the set $\trzy =\{0,1,\bot\}$ of truth values,
interpreted as {\em true}, {\em false\/} and {\em undefined},
respectively.  The subset of $\trzy $ consisting of $0$ and $1$ will
be denoted $\dwa.$

It follows from the discussion above that we are going to look for
conditionals in the set $\PC=\trzy ^{\Omega^+}$ of three-valued
functions $c$ from the set $\Omega^+$ of finite nonempty sequences of
atomic events from $\Omega$ into $\trzy .$ We will call such functions
{\em pre-conditionals}, since to deserve the name of conditionals they
must obey some additional requirements.

Sometimes it is convenient to represent such objects in two other, slightly
different, yet equivalent forms: 

\begin{itemize}
\item The second representation are length-preserving mappings
$c_+:\Omega^+\to\trzy ^+$ such that $c_+(v)$ is a prefix of $c_+(vw).$
The set of all such mappings will be denoted $\PC_+.$
\item The third representation are mappings
$c_\infty:\Omega^\infty\to\trzy ^\infty$ such that if
$w,v\in\Omega^\infty$ have a common prefix of length $n,$ then
$c_\infty(w)$ and $c_\infty(v)$ have a common prefix of length $n,$
too.  The set of all such mappings will be denoted $\PC_\infty.$
\end{itemize}

On the set $\Omega^+\cup\Omega^\infty$ one has the natural partial
order relation of being a prefix. Suprema of sets in this partial
order are denoted by $\bigsqcup.$

In general, $c,\ c_+$ and $c_\infty$ denote always three
representations of the same pre-conditional, and the subscript (or its
lack) indicates what representation we take at the moment, and we
choose it according to what is most convenient. The three
representations $c,\ c_+$ and $c_\infty$ are related by the equalities

\begin{equation}\label{reprezentacje}
\begin{split}
c(\omega_1\dots \omega_n)&=\text{last-letter-of}(c_+(\omega_1\dots
\omega_n)),\\
c(\omega_1\dots \omega_n)&=\text{$n$th-letter-of}(c_\infty(\omega_1\dots
\omega_n\dots)),\\
c_+(\omega_1\dots \omega_n)&=c(\omega_1)c(\omega_1\omega_2)\dots
c(\omega_1\dots \omega_n),\\ 
c_+(\omega_1\dots \omega_n)&=\text{first-$n$-letters-of}
(c_\infty(\omega_1\dots\omega_n\dots)),\\ 
c_\infty(\omega_1\dots \omega_n\dots)&=c(\omega_1)c(\omega_1\omega_2)\dots
c(\omega_1\dots \omega_n)\dots,\\ 
c_\infty(\omega_1\dots \omega_n\dots)&=\bigsqcup\{c_+(\omega_1\dots
\omega_n)~/~n=1,2\dots\}.
\end{split}
\end{equation}

Even though we are on a rather preliminary level of our construction, we
can address the general question of defining connectives among
pre-conditionals already now. In our setting such a connective is indeed
a function from some power of the space of pre-conditionals into itself.
However, to fulfill the requirement that a connective should depend
solely on the outcomes of its arguments (this property is called {\em
extensionality\/} in the logic literature), and that it should refer to
the history, only, the following additional condition must be met.

For any connective $\alpha:\PC_+^n\to \PC_+$
and any $\varphi_1,\dots,\varphi_n,\varphi'_1,\dots,\varphi'_n\in
\PC_+,$ $v,w\in\Omega^+$ satisfying $\varphi_i(w)=\varphi'_i(v)$ for
$i=1,\dots,n$ holds

\[\alpha(\varphi_1,\dots,\varphi_n)(w)=
\alpha(\varphi'_1,\dots,\varphi'_n)(v).\]

Note that we permit strong dependence on the history: we do not
require the connective to depend just on the present values of its
arguments, we allow it to depend on their whole histories. However, if
a particular connective $\alpha$ meets the former, stronger
requirement, whose formal statement can be obtained from the above
condition by replacing $\PC_+$ by $\PC$ everywhere it occurs, we call
it a {\em present tense connective}.

Connectives which are not present tense will be called {\em past
tense}. Any $n$-ary present tense connective of pre-conditionals is
fully characterised by a mapping $\trzy ^n\to \trzy.$ Note that any
connective $\alpha,$ not necessarily present tense one, can be
completely specified by a mapping $\bigcup_{t>0}\underbrace{\trzy
^t\times\dots\times\trzy ^t}_{\text{$n$ times}}\to \trzy .$

Just like their connectives, pre-conditionals can be present tense,
too. A pre-conditional $c:\Omega^+\to\trzy $ is called {\em present
tense\/} iff $c(v)=c(w)$ holds whenever
$\text{last-letter-of}(v)=\text{last-letter-of}(w).$ So indeed a
present tense pre-conditional is completely determined by a function
$\Omega\to\trzy .$

\subsection{The formalisms}

Our intention is to distinguish conditionals among pre-conditionals.
Therefore, in order to deal with them, we need a formalism aimed at
dealing with sequences of symbols from a finite alphabet. There are
many candidates of this kind, including regular expressions and their
subclasses, grammars of various kinds, deterministic or
nondeterministic automata, temporal logics, first order logic and
higher order logics.

Our choice, which will be carefully motivated later on, is to use
three-valued counterparts of a certain particular class of finite
automata and of past tense temporal logic. When the probabilities come
into play conditional events of a fixed probability space are
represented by Markov chains.

We introduce here briefly the main formalisms used throughout this
paper: temporal logic, Moore machines and Markov chains.

\subsection{Temporal logic}

Let us first define {\em temporal logic of linear discrete past time},
called $\TL.$ We follow the exposition in \cite{Emerson}, tailoring
the definitions somewhat towards our particular needs.

The formulas are built up from the set $\E$ (the same set of basic
events as before), interpreted as propositional variables here, and
are closed under the following formula formation rules:

\begin{enumerate}
\item Every $a\in \E$ is a formula of temporal logic.
\item If $\varphi,\psi\in \TL,$ then their boolean combinations
$\varphi\lor \psi$ $\lnot\varphi$ are in $\TL.$ The other Boolean
connectives: $\land,\to,\lra,\dots$ can be defined in terms of $\lnot$
and $\lor,$ as usual.

\item If $\varphi,\psi\in \TL,$ then their past tense temporal
combinations $\PREV\varphi$ and $\varphi\Since\psi$ are in $\TL,$
where $\PREV\varphi$ is spelled ``previously $\varphi.$''
\end{enumerate}

A model of temporal logic is a sequence $\M=s_0,s_1,\dots,s_n$ of states, each
state being a function from $\E$ (the same set of basic events as before) to
the boolean values $\{0,1\}.$ Note that a state can be therefore understood as
an atomic event from $\Omega,$ and $\M$ can be thought of as a word from
$\Omega^+.$ To be explicit we declare that the states of
$\M$ are ordered by $\leq.$ Rather than using the indices of states to denote
their order, we simply write $s\leq t$ to denote that a state $t$ comes later
than, or is equal to, a state $s;$ similarly $s+1$ denotes the successor state
of $s.$ We adopt the convention that, unless explicitly indicated otherwise, a
model is always of length $n+1,$ and thus $n$ is always the last state of a
model.

For every state $s$ of $\M$ we define inductively what it means that a
formula $\varphi\in\TL$ is satisfied in the state $s$ of $\M,$
symbolically $\M,s\models\varphi.$

\begin{enumerate}
\item $\M,s\models a$ iff $s(a)=1$
\item
\begin{align*}
\M,s\models\lnot\varphi&:\iff\ \M,s\not\models\varphi,\\
\M,s\models\varphi\lor\psi&:\iff
\M,s\models\varphi\ \text{or}\ \M,s\models\psi.
\end{align*}

\item
\begin{align*}
\M,s\models\PREV\varphi&:\iff s>0\ \text{and}\ \M,s-1\models\varphi;\\
\M,s\models\varphi\Since\psi&:\iff(\exists t\leq s)(\M,t\models\psi\
\text{and}\ (\forall t< w\leq s)M,w\models\varphi).
\end{align*}
\end{enumerate}

The syntactic abbreviations $\PBOX\varphi$ and $\PDIA\varphi$ are of
common use in $\TL.$ They are defined by $\PDIA\varphi\equiv\FALSE
\Since\varphi$ and $\PBOX\varphi\equiv\lnot\PDIA\lnot\varphi.$ The
first of them is spelled ``once $\varphi$'' and the latter ``always in
the past $\varphi$''.
 
Their semantics is then equivalent to
\begin{align*}
\M,s\models\PBOX\varphi&:\iff(\forall t\leq s)\M,t\models\varphi;\\
\M,s\models\PDIA\varphi&:\iff(\exists t\leq s) \M,t\models\varphi.
\end{align*}

Using the given temporal and boolean connectives, one can write down
quite complex formulae describing temporal properties of models
$\M,s.$ We will see several such examples in this paper, and even more
can be found in  \cite{cea}.

\subsection{Moore machines}\label{MM}

In this section we follow \cite{HU}, tailoring the definitions, again,
towards our needs.

A {\em deterministic finite automaton\/} is a five-tuple
$\A=(Q,\Omega,\delta,q_0,T),$ where $Q$ is its set of states, $\Omega$
(the same set of atomic events as before) is the input alphabet,
$q_0\in Q$ is the initial state and $\delta: Q \times\Omega\to Q$ is
the transition function.  $T\subseteq Q$ is the set of accepting
states.

We picture $\A$ as a labelled directed graph, whose vertices are
elements of $Q,$ a the function $\delta$ is represented by directed
edges labelled by elements of $\Omega$: the edge labelled by
$\omega\in\Omega$ from $q\in Q$ leads to $\delta(q,\omega).$ The
initial state is typically indicated by an unlabelled edge ``from
nowhere'' to this state.

As the letters of the input word $w\in\Omega^+$ come in one after
another, we walk in the graph, always choosing the edge labelled by
the letter we receive. What we do with the word depends on the state
we are in upon reaching  the end of the word. If it is in $T,$ the
automaton accepts the input, otherwise it rejects it.

Formally, to describe the computation of $\A$ we extend $\delta$ to a
function $\hat{\delta}: Q \times\Omega^+\to Q$ in the
following way: 

\[\hat{\delta}(q,w)=\begin{cases}
\delta(q,w)&\text{if $|w|=1$}\\
\delta(\hat{\delta}(q,v),\omega)&\text{if $w=v\omega.$}
                    \end{cases}
\]

$L(\A)\subseteq\Omega^+$ is the set of words accepted by $\A.$

A {\em Moore machine $\A$} is a six-tuple
$\A=(Q,\Omega,\Delta,\delta,h,q_0),$ where $(Q,\Omega,\delta,q_0)$ is
a deterministic finite automaton but the set of accepting states,
$\Delta$ is a finite output alphabet and $h$ is the output function
$Q\to\Delta.$ In addition to what $\A$ does as a finite automaton, at
each step it reports to the outside world the value $h(q)$ of the
state $q$ in which it is at the moment.  Drawing a Moore machine we
indicate $h$ by labelling the states of its underlying finite
automaton by their values under $h.$ In addition, we almost always
make certain graphical simplifications: we merge all the transitions
joining the same pair of states into a single transition, labelled by
the union (evaluated in $\Sigma$) of all the labels. Sometimes we go
even farther and drop the label altogether from one transition, which
means that all the remaining input letters follow this transition.

Formally, a Moore machine computes a function $f_\A:\Omega^+\to
\Delta^+$ defined by

\[
f_\A(\omega_1\omega_2\dots\omega_n)
=h(\hat{\delta}(q_0,\omega_1))h(\hat{\delta}(q_0,\omega_1\omega_2))\dots
h(\hat{\delta}(q_0,\omega_1\omega_2\dots\omega_n))
\]

(note that $|f_\A(\omega_1\omega_2\dots\omega_n)|=n,$ as desired), and
a function $g_\A:\Omega^\infty\to \Delta^\infty$ defined by

\[g_\A(\omega_1\omega_2\dots)=
\bigsqcup\{f_\A(\omega_1\omega_2\dots\omega_n)~/~n=1,2,\dots\}.\]

We will be interested in Moore machines which compute $\trzy $-valued
functions.  This amounts to partitioning the state set $Q$ of $\A$
into three subsets $T,F,B,$ which we often make into parts of the
machine.  If we do so, we call the states in $T$ the {\em accepting
states\/} and the states in $F$ the {\em rejecting states}.  There
will be no special name for the states in $B.$

A Moore machine $\A$ is called {\em counter-free\/} if there is no
word $w\in\Omega^+$ and no states $q_1,q_2,\dots,q_s,\ s>1,$ such that
$\hat\delta(q_1,w)=q_2,\dots,\hat\delta(q_{s-1},w)=
q_s,\hat\delta(q_s,w)=q_1.$

\subsection{Markov chains}

For us, Markov chains are a synonym of {\em Markov chains with
stationary transitions and finite state space.}

Formally, given a finite set $I$ of {\em states} and a fixed
function $p:I\times I\to[0,1]$ satisfying 

\begin{equation}\label{stoch}
(\forall i\in I)\qquad\sum_{j\in I}p(i,j)=1,
\end{equation}
the {\em Markov chain\/} with state space $I$ and transitions $p$ is a
sequence $\X=X_0,X_1,\dots$ of random variables $X_n:W\to I$, such
that

\begin{equation}\label{M1}\Pr(X_{n+1}=j|X_n=i)=p(i,j).\end{equation}

The standard result of probability theory is that there exists a probability
triple $(W,\fM,\Pr)$ and a sequence $\X$ such that \eqref{M1} is
satisfied. $W$ is indeed the space of infinite sequences of ordered pairs
of elements from $I,$ and $\Pr$ is a certain product measure on this set.

One can arrange the values $p(i,j)$ in a matrix $\Pi=(p(i,j);i,j\in
I).$ Of course, $p(i,j)\geq 0$ and $\sum_{j\in I}p(i,j)=1$ for every
$i.$ Every real square matrix $\Pi$ satisfying these conditions is
called {\em stochastic.}  Likewise, the initial distribution of $\X$
is that of $X_0,$ which can be conveniently represented by a vector
$\Xi_0=(p(i); i\in I).$ Its choice is independent from the function
$p(i,j).$

It is often very convenient to represent Markov chains by matrices,
since many manipulations on Markov chains correspond to natural
algebraic operations performed on the matrices.

For our purposes, it is convenient to imagine the Markov chain $\X$ in
another, equivalent form: Let $K_I$ be the complete directed graph on
the vertex set $I.$ First we randomly choose the starting vertex in
$I,$ according to the initial distribution.  Next, we start walking in
$K_I;$ at each step, if we are in the vertex $i,$ we choose the edge
$(i,j)$ to follow with probability $p(i,j).$ If we define
$X_n=(\text{the vertex in which we are after $n$ steps}),$ then $X_n$
is indeed the same $X_n$ as in \eqref{M1}.

So we will be able to {\em draw\/} Markov chains.  Doing so, we will
often omit edges $(i,j)$ with $p(i,j)=0.$

\paragraph{Classification of states}

For two states $i,j$ of a Markov chain $\X$ with transition
probabilities $p$ we say that $i$ {\em communicates\/} with $j$ iff
there is a nonzero probability of eventually getting from $i$ to $j.$
Equivalently, it means that there is a sequence
$i=i_1,i_2,\dots,i_n=j$ of states such that $p(i_k,i_{k+1})>0$ for
$k=1,\dots,n-1.$ The reflexive relation of mutual communication (i.e.,
that $i$ communicates with $j$ and $j$ communicates with $i$ or $i=j$)
is an equivalence relation on $I.$ Class $[i]$ communicates with class
$[j]$ iff $i$ communicates with $j.$

The relation of communication is a partial ordering relation on
classes. The minimal elements in this partial ordering are called {\em
ergodic sets}, and nonminimal elements are called {\em transient
sets}. The elements of ergodic and transient sets are called ergodic
and transient states, respectively.

A Markov chain all whose ergodic sets are one-element is called {\em
absorbing}, and its ergodic states are called absorbing.

For ergodic sets one can be further define their {\em period.} Period
of an ergodic state $i$ is the gcd of all the numbers $p$ such that
there is a sequence $i=i_1,i_2,\dots,i_p=i$ of states such that
$p(i_k,i_{k+1})>0$ for $k=1,\dots,p-1.$ It can be shown that period is
a class property, i.e., all states in one ergodic class have the same
period. 

An ergodic set is called {\em aperiodic\/} iff its period is $1.$
Equivalently, it means that for every two $i,j$ in this set and all
sufficiently large $n$ there exists a sequence $i=i_1,i_2,\dots,i_n=j$
of states such that $p(i_k,i_{k+1})>0$ for $k=1,\dots,n-1.$

Every periodic class $C$ of period $p>1$ can be partitioned into $p$
periodic sub-classes $C_1,\dots,C_p$ such that $\Pr(X_{n+1}\in
C_{k+1\pmod p}|X_{n}\in C_{k\pmod p})=1$ for all $k.$

\section{Constructing conditionals}

We make a terminological distinction. If we speak about a {\em
conditional object}, we do not assume any probability space structure
imposed on $\Omega.$ When we have such structure
$(\Omega,\Sigma,\Pr),$ we speak about a {\em conditional event},
instead.

\subsection{Conditional objects}

First of all, let us note that any $\TL$ formula can be understood as
a definition of a pre-conditional from $\PC,$ which is indeed
$\dwa$-valued. Indeed, states of any model of temporal logic can be
interpreted as elements of $\Omega,$ and the whole model is thus an
element of $\Omega^+.$ The value the pre-conditional assigns to 
model $\M$ is $1$ if $\M,n\models\varphi$ and $0$ otherwise.

We construct a three-valued extension $\TT$ of $\TL$ as the set of
all pairs $(\varphi|\psi)$ of formulas from $\TL.$ The operator
$(\cdot|\cdot)$ can be understood as a {\em present tense
connective\/} of pre-conditionals, and, since formulas of $\TL$ are
$\dwa$-valued, it is sufficient to define its action as follows:

\[\begin{array}{|c|c|c|c|} \hline
\multicolumn{4}{|c|}{(x | y)}\\ \hline\hline x\diagdown y &0 &1 &\bot\\
\hline 0 &\bot &0 &~\\ \hline 1 &\bot &1 &~\\ \hline \bot &~ &~
&~\\ \hline
\end{array}
\]

\begin{definition}
A {\em conditional object of type 1\/} is a pre-conditional $c\in
\PC,$ definable in $\TT.$ The set of such conditional objects is denoted
$\CC.$
\end{definition}

\begin{definition} A {\em conditional object of type 2\/} is a pre-conditional
$c_+\in \PC_+,$ such that $c_+$ is computable by a $\trzy $-valued
counter-free Moore machine. The set of such conditional objects is denoted
$\CC_+.$
\end{definition}

\begin{definition} A {\em conditional object of type 3\/} is a pre-conditional
$c_\infty\in \PC_\infty,$ such that $c_\infty$ is computable by a $\trzy
$-valued counter-free Moore machine. The set of such conditional objects is
denoted $\CC_\infty.$
\end{definition}

The following proposition says that the conditional objects of types
1, 2 and 3 are identical up to the way of representing
pre-conditionals.

\begin{theorem}\label{tfae}
\begin{align*}
\CC_+&=\{c_+\in \PC_+~/~c\in \CC\},\\
\CC_\infty&=\{c_\infty\in\PC_\infty~/~c\in\CC\},\\
\CC&=\{c\in \PC~/~c_\infty\in \CC_\infty\}.
\end{align*}
\end{theorem}
\begin{proof}
The equalities $\CC_+=\{c_+\in \PC_+~/~c_\infty\in \CC_\infty\}$ and
$\CC_\infty=\{c_\infty\in\PC_\infty~/~c_+\in\CC_+\}$ are obvious.
What remains to be proven are $\CC_+=\{c_+\in \PC_+~/~c\in \CC\}$ and
$\CC=\{c\in\PC~/~c_+\in\CC_+\}$

It is well-known \cite{Emerson} that propositional temporal logic of
past tense and (finite) deterministic automata are of equal expressive
power, i.e., in our terminology, the sets of $\dwa$-valued
pre-conditionals from $\PC$ definable in $\TL$ and computable by
deterministic finite automata are equal. Indeed the translations
between temporal logic and automata are effective.

We start with the first equality. Let $c$ be defined by a $\TT$
formula $(\varphi|\psi).$ Let $\A=(Q_\A,\Omega,\delta_\A,q_\A,T_\A)$
and $\B=(Q_\B,\Omega,\delta_\B,q_\B,T_\B)$ be deterministic finite
automata, computing the functions $\Omega^+\to\dwa$ defined by
$\varphi$ and $\psi,$ respectively.

Consider the Moore machine $(\A|\B)=(Q_\A\times
Q_\B,\Omega,\trzy,\delta,h,(q_\A,q_\B)),$ where
\begin{align*}
\delta((p,q),\omega)&=(\delta_\A(p,\omega),\delta_\B(q,\omega)),\\
h((p,q))&=\begin{cases}1&\text{if $p\in T_A$ and $q\in T_\B$},\\
                     0&\text{if $p\notin T_A$ and $q\in T_\B$},\\
                     \bot&\text{otherwise.}
        \end{cases}
\end{align*}

It is immediate to see that $(\A|\B)$ computes exactly
$(\varphi|\psi)_+.$

To prove the second equality, let $\A=(Q,\Omega,\trzy,\delta,h,q_0)$
be a Moore machine computing $c_+.$ We construct two deterministic
finite automata $\A_1=(Q,\Omega,\delta,q_0,\vec{h}^{-1}(\{1\}))$ and
$\A_2=(Q,\Omega,\delta,q_0,\vec{h}^{-1}(\{0,1\})$ from $\A,$ where
$\vec{h}^{-1}$ stands for the co-image under $h.$ Now let $\varphi_1$
and $\varphi_2$ be $\TL$ formulae corresponding to $\A_1$ and $\A_2,$
respectively. 

It is again immediate to see that $(\varphi_1|\varphi_2)$ defines
exactly the conditional in $\CC$ computed in $\CC_+$ by $\A.$
\end{proof}

Consequently, we can freely choose between the three available
representations of conditional objects. Doing so, we regard $\TT$ to
be the {\em logic of conditional objects}, while Moore machines
represent their {\em machine representation}. All these
representations are equivalent, thanks to Theorem~\ref{tfae}.

The classes $\CC,\ \CC_+$ and $\CC_\infty$ represent the {\em
semantics\/} of conditional objects, and again we can freely choose
the particular kind of semantical objects, thanks to
\eqref{reprezentacje}.

As an example, the simple conditional $(a|b)\in \TT$ is computed by
the following Moore machine.

\begin{figure}[!hbtp]

\[\UseTips
\xymatrix @C=15mm @R=10mm
{
{}\save[]+<5mm,-5mm>*{}\ar[rd]\restore&&*+++[o][F]{1}
\ar@(ul,ur)[]^{{ba}}
\ar[dd]^{{ba^\C}}
\ar@/_9mm/[dl]_{{b^\C}}\\
&*+++[o][F]{\bot}
\ar@(dl,l)[]^{{b^\C}}
\ar[ur]_{{ba}}
\ar@/_9mm/[dr]_{{ba^\C}}\\
&&*+++[o][F]{0}
\ar@(ld,dr)[]_{{ba^\C}}
\ar@/_9mm/[uu]_{{ba}}
\ar[ul]_{{b^\C}}
}
\]
\caption{Moore machine representing  conditional object
$(a|b).$}\label{f1}
\end{figure}
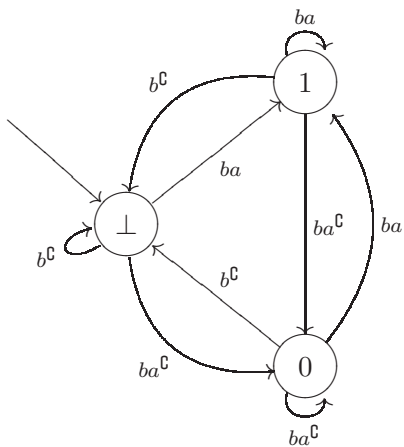

The above Moore machine, as it is easily seen, acts exactly according to the
rule ``ignore $b^\C$'s, decide depending on the truth status of $a$ when $b$
appears''. So indeed it represents the repetitions of the experiment for
$(a|b)$ according to the ``bridge'' repetition rule {\em start history anew}.

\subsection{Conditional events}

We will be using the name {\em conditional events\/} to refer to
conditionals considered with a probability space in the background.

Let $(\Omega,\PP(\Omega),\Pr)$ be a probability space.

\begin{definition}[Conditional event] Let $c\in \CC$ be a conditional object
over $\Omega.$ Suppose $\Omega$ is endowed with a probability space
structure $(\Omega,\Sigma,\Pr).$ With $c$ we associate the sequence
$\Y=\Y(c)=Y_1,Y_2,\dots$ of random variables $\Omega^\infty\to\trzy ,$
defined by the formula

\begin{equation}\label{e1}
Y_n(w)=\text{$n$-th-letter-of}(c_\infty(w)),
\end{equation}

where $\Omega^\infty$ is considered with the product probability structure.

We call $\Y$ the {\em conditional event\/} associated with $c,$ and denote it
$\ce{c},$ while $Y_n$ is then denoted $\ce{c}_n.$ Note that we do not include
the probability space in the notation. It will be always clear what
$(\Omega,\PP(\Omega),\Pr)$ is.
\end{definition}

In particular, $\Pr(\ce{c}_n=\true)$ is the probability that at time
$n$ the conditional is true, $\Pr(\ce{c}_n=\false)$ is the probability
that at time $n$ the conditional is false, and $\Pr(\ce{c}=\bot)$ is
the probability that at time $n$ the conditional is undefined.

\begin{definition}[Probability of conditional events]~\\
We define the {\em asymptotic probability at time $n$\/} of a
conditional $c$ by the formula

\begin{equation}\label{e2}
\pr_n(c)=\dfrac{\Pr(\ce{c}_n=1)}
{\Pr(\ce{c}_n=0\ \text{or}\ 1)}.
\end{equation}

If the denominator is $0,$ $\pr_n(c)$ is undefined.

The {\em asymptotic probability\/} of $c$ is

\begin{equation}\label{e3.5}
\Pr(c)=\lim_{n\to\infty}\pr_n(c),
\end{equation}

provided that $\pr_n(c)$ is defined for all sufficiently large $n$ and
the limit exists.

We will regard $\ce{c}$ as {\em probabilistic semantics\/} of $c.$

If $\varphi\in\TL$ then we write $\Pr(\varphi)$ for $\Pr((\varphi|\TRUE)).$
\end{definition}

It is perhaps reasonable to explain why we want the conditional event
and its probability to be defined in this way.  The main motivation is
that we want the conditional event and its probability to be natural
and intuitive. And we achieve this by using the recipe of de Finetti,
which in our case materializes in the above definitions.

\section{Underlying Markov chains, Bayes' Formula and classification of
conditional events} 
\label{classifying}

\subsection{Underlying Markov chains}

Let $c$ be a conditional object and let $\A=(Q,\Omega,\delta,\trzy,h,q_0)$ be
a counter-free Moore machine which computes $c_\infty.$

We define a Markov chain $\X=\X(\A)$ by taking the set of states of
$\X$ to be the set $Q$ of states of $\A,$ and the transition function
$p$ to be defined by

\begin{equation*}
p(q,q')=\sum_{\substack{\omega\in\Omega\\\delta(q,\omega)=q'}}\Pr(\{\omega\}).
\end{equation*}

Indeed, for every $q$ we have

\begin{equation*}
\sum_{q'}p(q,q')
=\sum_{q'}\sum_{\substack{\omega\in\Omega\\
\delta(q,\omega)=q'}}\Pr(\{\omega\}) 
=\sum_{\omega\in\Omega}\Pr(\{\omega\}) =1,
\end{equation*}

which means that the function $p$ satisfies \eqref{stoch}, which is
the criterion for being a transition probability function of a Markov
chain.  The initial probability distribution is defined by

\[
p(q)=
\begin{cases}1&\text{if $q=q_0,$ the initial state of $\A,$}\\
             0&\text{otherwise.}
\end{cases}
\]

Therefore we have indeed converted $\A$ into a Markov chain $\X.$

In the pictorial representation of the conversion process is much
simpler: we take the drawing of $\A,$ and replace all the letters from
$\Omega$ marking transitions by their probabilities according to $\Pr,$
and then contract multiple transitions between the same states into a
single one, summing up their probabilities.

\begin{theorem}\label{transient-aperiodic} $\X$ is a Markov chain in
which only transient and aperiodic states exist.
\end{theorem}

\begin{proof} Suppose $\X$ has a periodic set $C$ of period $p>1,$ and
$C_1,\dots,C_p$ its division into periodic subclasses. Let $\omega\in
\Omega$ be any atomic event with $\Pr(\{\omega\})>0.$ Let $q\in C_1.$
Since $\Pr(X_{n+1}\in C_{k+1\pmod p}|X_{n}\in C_{k\pmod p})=1$ for all
$k,$ it follows that $\delta^1(q,\omega)=\delta(q,\omega)\in C_{2\pmod
p},$ and likewise
$\delta^{k+1}(q,\omega)=\delta(\delta^k(q,\omega))\in C_{k+1\pmod p}$
for $k\geq 1.$

However, $C$ is finite, so there must be $s\neq t$ such that
$\delta^s(q,\omega)=\delta^t(\omega).$

The sequence 

\[\delta^s(q,\omega),\delta^{s+1}(q,\omega),\dots,\delta^t(q,\omega)
=\delta^s(q,\omega)\] 

thus violates the assumption that $\A$ is counter-free.\end{proof}

The next corollary follows by the classical result about finite Markov chains.

\begin{corollary}\label{limits}
For every state $i$ of $\X,$ the limit $\lim_{n\to\infty}\Pr(X_n=i)$
exists.
\end{corollary}

Using $h:Q\to\trzy,$ the {\em acceptance mapping of $\A,$} we get

\begin{theorem}
\(\label{Y=h(X)}\ce{c}=h(\X).\)\qed
\end{theorem}

Note that $\ce{c}$ defined above need not be a Markov chain
itself, but it is a simple projection of a Markov chain, extracting
all the invariant information.  Of course, it will be typically very
beneficial to work most of the time with $\X,$ having the whole theory
of Markov chains as a tool-set, and only then to move to $\ce{c}.$

Let us examine the previously given definition of $(a|b)$ to see what its
probability is.

The Markov chain looks as follows:

\begin{figure}[!hbtp]

\[\UseTips
\xymatrix @C=15mm @R=10mm
{
{}\save[]+<5mm,-5mm>*{}\ar[rd]\restore
&&*+++[o][F]{1}
\ar@(ul,ur)[]^{{\Pr(ba)}}
\ar[dd]^{{\Pr(ba^\C)}}
\ar@/_9mm/[dl]_{{\Pr(b^\C)}}\\
&*+++[o][F]{\bot}
\ar@(dl,l)[]^{{\Pr(b^\C)}}
\ar[ur]_{{\Pr(ba)}}
\ar@/_9mm/[dr]_{{\Pr(ba^\C)}}\\
&&*+++[o][F]{0}
\ar@(ld,dr)[]_{{\Pr(ba^\C)}}
\ar@/_14mm/[uu]_{{\Pr(ba)}}
\ar[ul]_{{\Pr(b^\C)}}
}
\]
\caption{Markov chain corresponding to the Moore machine on Fig.\
\ref{f1}.}\label{f2}
\end{figure}
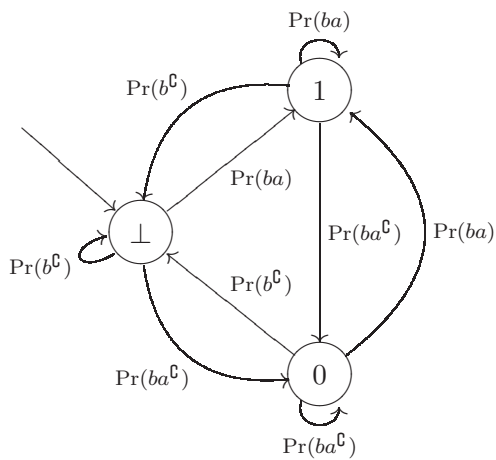

where the initial distribution assumes probability 1 given to the
state pointed to by the arrow ``from nowhere''.  

It is easy to check that $\Pr((a|b))=\Pr(ba)/\Pr(b),$ provided that
$\Pr(b)>0.$ Indeed, for every $n$ holds $\Pr(\ce{(a|b)}_n=1)=\Pr(ba)$
and $\Pr(\ce{(a|b)}_n=0)=\Pr(ba^\C),$ so $\Pr(\ce{(a|b)}_n=0\
\text{or}\ 1)=\Pr(ba)+\Pr(ba^\C)=\Pr(b).$ It is so because, no matter
in which state we are, these are the probabilities of getting to $1$
and $0$ in the next step, respectively. This evaluation will follow from
Bayes' Formula below, too.

\subsection{Bayes' Formula}
First of all, let us note that for each $\star\in\trzy$ the limit
$\lim_{n\to\infty}\pr(\ce{c}_n=\star)$ exists, since, for any choice
of a Moore machine $\A$ computing $c_+$ and assuming $\X=\X(\A),$
$\pr(\ce{c}_n=\star)$ is a sum of $\pr(X_n=i)$ over all states $i$ of
$\X$ with $h(i)=\star,$ and the latter probabilities converge by
Corollary \ref{limits}.

A conditional event is called {\em regular\/} iff
$\lim_{n\to\infty}\Pr(\ce{c}_n=0\ \text{or}\ 1)>0.$ In particular, for
regular conditionals the limit in \eqref{e3.5} always exists and is
equal to

\begin{equation*}
\dfrac{\lim_{n\to\infty}\Pr(\ce{c}_n=1)}
{\lim_{n\to\infty}\Pr(\ce{c}_n=0\ \text{or}\ 1)}.
\end{equation*}

Turning to the logical representation of conditionals,  we have thus

\begin{theorem}[Bayes' Formula]\label{Bayes}
For $(\varphi|\psi)\in\TT$ 
\[\Pr((\varphi|\psi))=\frac{\Pr(\varphi\land\psi)}{\Pr(\psi)}\]

whenever the right-hand-side above is well-defined.\qed
\end{theorem}

Note that Bayes' Formula has been expected by de Finetti for the frequency
based conditionals.

\subsection{Classifying conditional events}

It is interesting to consider the conditionals $c$ for which
$\lim_{n\to\infty}\Pr(\ce{c}_n=0\ \text{or}\ 1)=0.$ We can distinguish
two types of such conditional events: those for which $\Pr(\ce{c}_n=0\
\text{or}\ 1)$ is identically $0$ for infinitely many $n,$ and those
for which it is nonzero for all but finitely many $n$. The former will
be called {\em degenerate}, the latter {\em strange}. We call {\em
strictly degenerate\/} those degenerate events, for which
$\Pr(\ce{c}_n=0\ \text{or}\ 1)$ for all but finitely many $n.$

The degenerate conditional events correspond to bets which infinitely
often cannot be resolved, because they are undefined, and strictly
degenerate events are those which are almost never defined.

Strange conditional events are more interesting. The Bayes' Formula is
senseless for them, so we have to use some ad hoc methods to see if
their asymptotic probability exists or not.

The first example shows that the sequence $\pr_n(c)$ can be
nonconvergent for strange $c.$

Consider $c_1=(a|\PBOX ((\PREV a\to a^\C)\land(\PREV a^\C\to
a)\land(\lnot\PREV \TRUE\to a ))),$ where $0<\Pr(a)<1.$ The long
temporal formula asserts that $a$ always follows $a^\C$ and $a^\C$
always follows $a,$ and at the beginning of the process ($n=1$), where
$\PREV \TRUE $ is false, $a$ holds.

It is easily verified that 

\[\pr_n(\ce{c_1})=\begin{cases}
                 0&\text{if $n$ is even},\\
                 1&\text{if $n$ is odd}.
           \end{cases}
\]

Thus the finite-time behaviour of this conditional is not
probabilistic---its truth value depends solely on the {\em age\/} of
the system. So for somebody expecting a pure game of chances its
behaviour must seem strange (and hence the name of this class of
conditional events).

Note that we have just discovered the next feature of conditionals
expected by de Finetti: nonconvergence of the limiting frequency when
probability of the `given' part tends to $0.$

However, again following de Finetti, if $(\varphi|\varphi)$ is
strange, its asymptotic probability is $1.$ E.g., $\Pr(\PBOX ((\PREV
a\to a^\C)\land(\PREV a^\C\to a)\land(\lnot\PREV \TRUE\to a ))|\PBOX
((\PREV a\to a^\C)\land(\PREV a^\C\to a)\land(\lnot\PREV \TRUE\to a
)))=1.$

Moreover, for $c_2=(a|\PBOX ((\PREV a\to  a^\C)\land(\PREV a^\C\to 
a)))$ we have 
\[\pr_n(\ce{c_2})=\begin{cases}
                 1-\Pr(a)&\text{if $n$ is even},\\
                 \Pr(a)&\text{if $n$ is odd}.
           \end{cases}
\]

Indeed, here the `given' part requires that $a$'a and $a^\C$'s
alternate, but does not specify what is the case at the beginning of
the process. So the probability of the whole conditional at odd times
is the probability that $a$ has happened at time 1, and at even times
it is the probability that $a$ has not happened at time 1.  Therefore,
when $\Pr(a)=1/2,$ $\Pr(c_2)$ exists and is $1/2.$ So asymptotic
probabilities which are neither $0$ nor $1$ are possible for strange
conditionals events.

At present, the question whether there it is decidable if a given
strange conditional event has an asymptotivc probability is open.
However, we believe that te answer is positive and offer it as our
cojecture.

\begin{conj}
The set of conditional events which have asymptotic probability is
decidable.  Moreover, for those events which have asymptotic
probability, its value is effectively computable. 
\end{conj}

\section{Connectives of conditionals}

\subsection{Present tense connectives}

Let us recall that present tense connectives are those, whose definition
in $\TT$ does not use temporal connectives, and therefore depends
on the present, only. Equivalently, an $n$-ary present tense connective is
completely characterised by a function $\trzy ^n\to\trzy .$ 

Here are several possible choices for the conjunction, which is always
defined as a pointwise application of the following $\trzy $ valued
functions. Above we display the notation for the corresponding kind of
conjunction.

\[
\conn{\land_{\SAC}}00001101\ \ \ \ 
\conn{\land_{\GNW}}00001\bot0\bot\ \ \ \
\conn{\land_{\Sch}}00\bot01\bot\bot\bot
\]
\[{\begin{array}{|c|c|}\hline 
\multicolumn{2}{|c|}{\sim x} \\\hline\hline x&\sim x\\\hline 0&1\\\hline
1&0\\\hline\bot&\bot\\\hline\end{array}}
\]
\[\conn{\lor_{\SAC}}01011101\ \ \ \ 
\conn{\lor_{\GNW}}0111\bot1\bot\bot\ \ \ \
\conn{\lor_{\Sch}}01\bot11\bot\bot\bot.
\]

They can be equivalently described by syntactical manipulations in $\TT.$
The reduction rules are as follows:

\begin{equation}\label{TLTL}
\begin{split}
(a|b)\land_\SAC(c|d)&=(abcd\lor abd^\C\lor cdb^\C|b\lor d)\\
(a|b)\land_\GNW(c|d)&=(abcd|a^\C d\lor c^\C d\lor abcd)\\
(a|b)\land_\Sch(c|d)&=(abcd|bd)\\
\sim (a|b)&=(a^\C|b)\\
(a|b)\lor_\SAC(c|d)&=(ab\lor cd|b\lor d)\\
(a|b)\lor_\GNW(c|d)&=(ab\lor cd|ab\lor cd \lor bd)\\
(a|b)\lor_\Sch(c|d)&=(ab\lor cd|bd).
\end{split}
\end{equation}

The first is based on the principle ``if any of the arguments becomes
defined, act!''.  A good example would be a quotation from \cite{c97}:

\begin{quote}\sl ``One of the most dramatic examples of the unrecognised
use of compound conditioning was the first military strategy of our
nation.   As the Colonialists waited for the British to attack, the
signal was `One if by land and two if by sea'.  This is the
conjunction of two conditionals with uncertainty!''
\end{quote}

Of course, if the above was understood as a conjunction of two
conditionals, the situation was crying for the use of $\land_\SAC,$
whose definition has been proposed independently by Schay, Adams and
Calabrese (the author of the quotation).

The conjunction $\land_{\GNW}$ represents a moderate approach, which
in case of an apparent evidence for $\false$ reports $\false,$ but
otherwise it prefers to report unknown in a case of any doubt. Note
that this conjunction is essentially the same as {\em lazy
evaluation},  known from programming languages. 

Finally, the conjunction $\land_\Sch$ is least defined, and acts
(classically) only if both arguments become defined. It corresponds to
the {\em strict evaluation}.

We have given an example for the use of $\land_\SAC.$ The uses of
$\land_\GNW$ and $\land_\Sch$ can be found in any computer program
executed in parallel, which uses either lazy or strict evaluation of
its logical conditions. And indeed both of them happily coexist in
many programming languages, in that one of them is the standard
choice, the programmer can however explicitly override the default and
choose the other evaluation strategy.

Let us mention that all the three systems above are in fact
well-known, classical so to say three-valued logics: 
$\langle\land_\GNW,\lor_\GNW,\sim\rangle$ is the  logic of
\L{}ukasiewicz, $\langle\land_\SAC,\lor_\SAC,\sim\rangle$ is the logic
of Soboci\'nski, and $\langle\land_\Sch,\lor_\Sch,\sim\rangle$ is the
logic of Bochvar.

\subsection{Past tense connectives} 

The following connective is tightly related to very close to the conjunction
of the {\em product space conditional event algebra\/} introduced in
\cite{g94}. Detailed discussion of embeddings of existing algebras of
conditional events into $\TT$ is included in the companion paper
\cite{cea}. Our new conjunction, denoted $\land^\star,$ is defined
precisely when at least one of its arguments is defined, so it
resembles $\land_\SAC$ in this respect, but instead of assigning the
other argument a default value when it is undefined, like $\SAC$ does,
it uses its most recent defined value, instead.  However, when the
other argument hasn't ever been defined, it is assumed to act like
$\FALSE.$

In the language of $\TT$ $(a|b)\land^\star(c|d)$ can be expressed by
\[((b^\C\Since (a\land b))\land(d^\C\Since (c\land
d))|b\lor d).\]

\subsection{Conclusion}

We believe that there is no reason to restrict our attention to any
particular choice of an operation extending the classical conjunction,
and call is {\em the conjunction of conditionals.} There are indeed
many reasonable such extensions, which correspond to different
intuitions and situations, they can coexist in a single formalism, and
any restriction in this respect necessarily narrows the applicability
of the formalism.

We believe that neither of the choices discussed in this paragraph is {\em
the\/} conjunction of conditionals. There are indeed many possible choices,
and all of them have their own merits.  In fact already the original system of
Schay consisted of five operations:  $\sim,\land_\SAC,\lor_\SAC,\land_\Sch$
and $\lor_\Sch.$ Moreover, he was aware that these operations still do not
make the algebra functionally complete (even in the narrowed sense, restricted
to defining only operations which are undefined for all undefined arguments).
And in order to remedy this he suggested to use one of several additional
operators, one of them being $\land_\GNW!$ So for him all those operations
could coexist in one system.

\section{Three prisoner's puzzle}

In order to demonstrate that our formalism allows for a precise
treatment of problems with conditioning and probabilities, let us
consider the following classical example of a probabilistic
``paradox''.  We will take this opportunity to highlight some of the
practical issues of modelling using $\TT$ and Moore machines
approach. Therefore our analysis will be very detailed.

\subsection{The puzzle}

The three prisoner's puzzle  \cite{p88} is the following: 

\begin{quote}
Three prisoners are sentenced for execution. One day before their
scheduled execution, prisoner $A$ learns that two of them have been
pardoned.  $A$ calculates a probability of $2/3$ for him being
pardoned.  Then he asks the Guard: ``Name me one of my fellows who
will be pardoned.  The Guard tells him, that $B$ will be pardoned.
Based on that information, $A$ recalculates the probability of being
pardoned as $1/2$, since now only one pardon remains for him and $C$
(the third prisoner) to share!  However, he could apply the same
argument if the Guard had named $C$.  Furthermore, he knew beforehand
that at least one of his fellows will be pardoned --- so what did he
gain (or lose) by the answer?
\end{quote}

The intuitive explanation is that after learning the Guard's testimony
$G(B)$ that $B$ will be pardoned, $A$ should revise the probability of
the event $P(A)$ (of him being pardoned) by computing $P(\PREV
P(A)|G(B))$, and the probability evaluation yields in this case
$2/3,$ as expected.

However, what he indeed calculated was $P(\PREV G(B)|P(A))$, assuming
effectively that the pardon had been given with equal probabilities to
all the pairs possible {\em after\/} Guard's testimony. This
probability turns out to be $1/2.$

\subsection{Probability tree model}

First we present a simple probability tree analysis of the paradox,
using the method which originates with Huygens \cite{huygens,shafer}
and is indeed almost as old as the mathematically rigorous probability
theory itself. We begin in the leftmost circle (before pardon), then
each of the three pardoned pairs leads us to three next circles,
indicating the situation after the pardon. Finally, we have all the
possible testimonies of the Guard. All edges originating from the same
circle are equiprobable. After Guard's testimony $G(B),$ only the two
top circles on the right are possible, and their probabilities are in
the proportion $2:1,$ the more probable one being the one in which $A$
is pardoned, while he is executed in the other one. So indeed even
after the testimony the probability that $A$ is pardoned remains
$2/3.$ 
\begin{figure}[!hbtp]
\[\UseTips
\xymatrix @C=20mm
{
&&*+++[o][F]{}
\\
&*+++[o][F]{}
\ar[ur]^{\textstyle{G(B)}}
&*+++[o][F]{}
\\
*+++[o][F]{}
\ar[ur]^{\textstyle{P(AB)}}
\ar[r]^(.7){\textstyle{P(BC)}}
\ar[dr]_{\textstyle{AC}}
&*+++[o][F]{}
\ar[ur]^{\textstyle{G(B)}}
\ar[dr]_{\textstyle{G(C)}}
\\
&*+++[o][F]{}
\ar[dr]_{\textstyle{G(C)}}
&*+++[o][F]{}
\\
&&*+++[o][F]{}
}
\]
\caption[]{Probability tree analysis of the three prisoner
puzzle.}
\label{f6}
\end{figure}
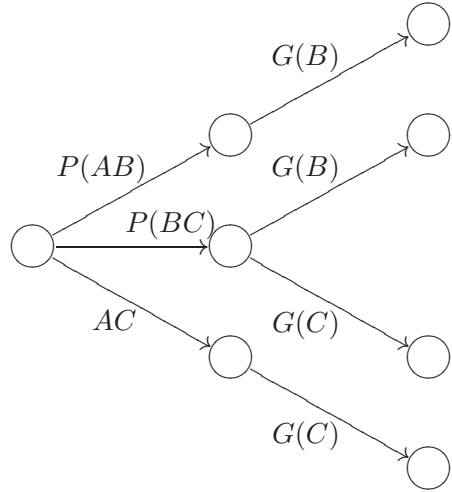

\subsection{$\TT$ and Moore machine models}

However, the tree shown above strongly resembles a Moore machine. And
indeed, we augment it with the necessary details below. The most
substantial change is that the Moore machine requires the same set of
atomic possibilities is given at each state, which determine the next
transition. Therefore:

\begin{itemize}
\item The Guard testifies something irrelevant while the court decides
the pardons, and the court decides something irrelevant while the Guard
testifies. This change is made invisible by our convention of
collapsing transitions and applying subsequently Boolean algebra
simplifications, except that
\item In cases when the Guard has no choice, we must replace the
existing transition label by the full event, because the Guard has
prescribed answer no matter whom he would like to name,
\item And except that we have to decide about transitions from the
states which are terminal in the tree model.  Because we believe that
after being pardoned nobody can be prosecuted again for the same
crime, and we do not believe in reincarnation, either, our choice is
to use self-loops in the terminal states, yielding a ``Russian
roulette'' model.
\end{itemize}

\begin{figure}[!hbtp]
\[\UseTips
\xymatrix @C=20mm
{
&&&*+++[o][F]{\textcolor{green}{1}}
\ar@{.>}@(ur,dr)[]
\\
&&*+++[o][F]{\textcolor{green}{\bot}}
\ar[ur]^{\textcolor{green}{\textstyle{\Omega}}}
&*+++[o][F]{\textcolor{green}{0}}
\ar@(ur,dr)@{.>}[]
\\
\ar@{.>}[r]
&*+++[o][F]{\textcolor{green}{\bot}}
\ar[ur]^{\textstyle{P(AB)}}
\ar[r]^(.67){\textstyle{P(BC)}}
\ar[dr]_{\textstyle{AC}}
&*+++[o][F]{\textcolor{green}{\bot}}
\ar[ur]_{\textstyle{G(B)}}
\ar[dr]^{\textstyle{G(C)}}\\
&&*+++[o][F]{\textcolor{green}{\bot}}
\ar[dr]_{\textcolor{green}{\textstyle{\Omega}}}
&*+++[o][F]{\textcolor{green}{\bot}}
\ar@{.>}@(ur,dr)[]
\\
&&&*+++[o][F]{\textcolor{green}{\bot}}
\ar@{.>}@(ur,dr)[]
}
\]
\caption[]{Probability tree analysis of the three prisoner puzzle with
{\color{green} extensions necessary to convert the diagram into a Moore
machine.}}
\label{f7}
\end{figure}
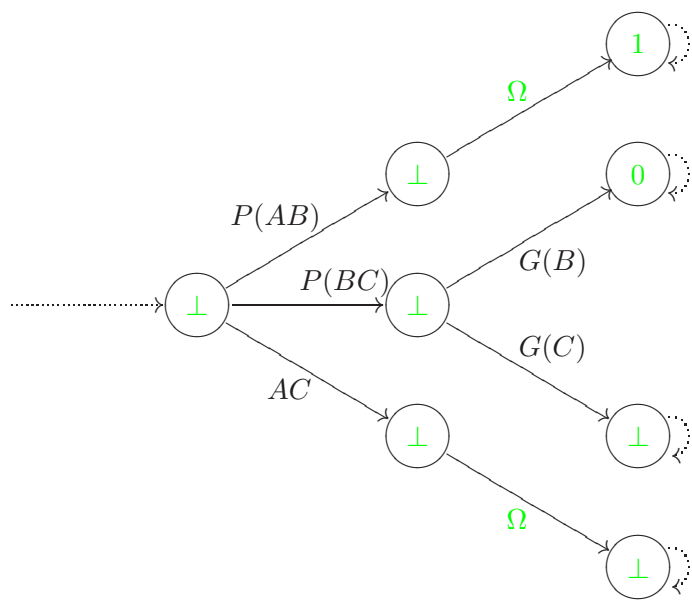

This provides a next piece of evidence that our definition of
conditional events is natural and close to intuitions.  In fact, one
can embed the whole probability tree model into the formalism of
Russian roulette Markov chains \cite{shafer}, and thus shows that our
model of conditionals extends the method of probability trees.

Next we attempt to model the same paradox syntactically in $\TT.$ The
construction of a correct $\TT$ representation is a little bit more
complicated than the formula $P(\PREV P(A)|G(B))$ we have suggested
previously, as this requires specifying the actions of the Guard,
whose probabilities are affected by the pardon decision. So we assume
that the Guard always tosses a coin. If he gets heads $(H)$, he tells
the alphabetically first name among those applicable, and in case of
tails $(T)$ the alphabetically last among them. This indicates the
need to consider the strategy followed by the Guard. And in fact, the
probabilities $A$ calculates depend on what he assumes about this
strategy. So indeed now the answers of the Guard are shorthands for
the combinations of the pardon decision and the coin toss outcome.
Therefore $G(B)$ is $(\PREV P(AB)\land(H\lor T))\lor (\PREV P(BC)\land
H).$

Moreover, we have to decide what should be modelled by the conditional
object, and what by the probability assignment, which turns the former
into a stochastic process.  The general rule is that the more of the
modelling is encoded in the probability assignment, the simpler the
conditional and its Moore machine are.  On the other hand, encoding
everything in the probability distribution is difficult and prone to
errors, as the example of the poor prisoner shows. An, needless to
say, a good model is one in which the proportions are just right. More
on that below.

So formally the conditional looks now as follows:

\begin{equation}\label{3pris}
\left(
(\PREV AB)
\lor
(\PREV AC)
\left|
((\PREV AB)\land (H\lor T))
\lor
((\PREV BC)\land H)
\right.\right),
\end{equation}

with $\E=\{P(AB),P(BC),AC,H,T\},$ where the events $P(AB),$ $P(BC)$
and $AC$ mutually exclusive and equiprobable, and similarly $H$ and
$T$ mutually exclusive and equiprobable. (Our construction will easily
handle non-equal probabilities, i.e., biased pardon decision and/or
biased coin, too.) So the set $\Omega$ of atomic events is $\{ AB\,H,
AB\,T, BC\,H, BC\,T,AC\,H, AC\,T\}, $ and these events are
equiprobable under our probability assignment. However, we will be
able to calculate the probability of \eqref{3pris} without the
equiprobability assumption, too.

Note that, e.g., assuming events $A,B$ and $C$ to be nonexclusive
individual pardon decisions of probability $1/3$ each, leads to more
complicated conditional expression, because a substantial amount of
coding effort must used just to ensure that always precisely two
prisoners are pardoned. This makes the Moore machine more complicated,
too. So this is certainly not a good model, because what can be easily
taken care of by the probability assignment is instead modelled by
logical methods. Such a model can be of course
correct,\footnote{Although unnecessary complications certainly
increase the risk of mistakes and make verification of the model
harder.} but good means for us more than just correct.

But if we attempt to draw the Moore machine of our conditional, we
discover that it is quite different from that on Fig.\ \ref{f7}.

\begin{figure}[!hbtp]
\[\UseTips
\xymatrix @R=18mm @C=9mm
{
&&*+++[o][F]{\bot}
&&*+++[o][F]{0}
\\
&*+++[o][F]{1}
\ar@(l,u)^(.63){AB(H\lor T)}
\ar[rrrr]|(.4){BC(H\lor T)}
\ar[ddrr]|(.44){AC(H\lor T)}
&&&&*+++[o][F]{1}
\ar[llllld]|{AB\,H}
\ar[lllu]|{AB\,T}
\ar[ul]|{BC\,H}
\ar[dr]|{BC\,T}
\ar[dd]|(.33){AC\,T}
\ar[lllldd]|(.375){AC\,H}[]
\\
*+++[o][F]{0}
&&&&&&*+++[o][F]{\bot}
\\
&*+++[o][F]{0}
\ar[uuur]|(.25){AB(H\lor T)}
\ar[urrrrr]|{BC(H\lor T)}
\ar@/_5ex/[rrrr]_{AC(H\lor T)}
&&*+++[o][F]{1}
&&*+++[o][F]{\bot}
\\
&&&&&
\ar@{.>}[r]
&*+++[o][F.]{\bot}
\ar@{.>}[uu]^(.7){BC}
\ar@{.>}@/_2.3em/[uuuullll]|(.19){AB}
\ar@{.>}[ul]^{AC}
}
\]
\caption[]{Moore machine corresponding to formula
\eqref{3pris}.}\label{f9}
\end{figure}
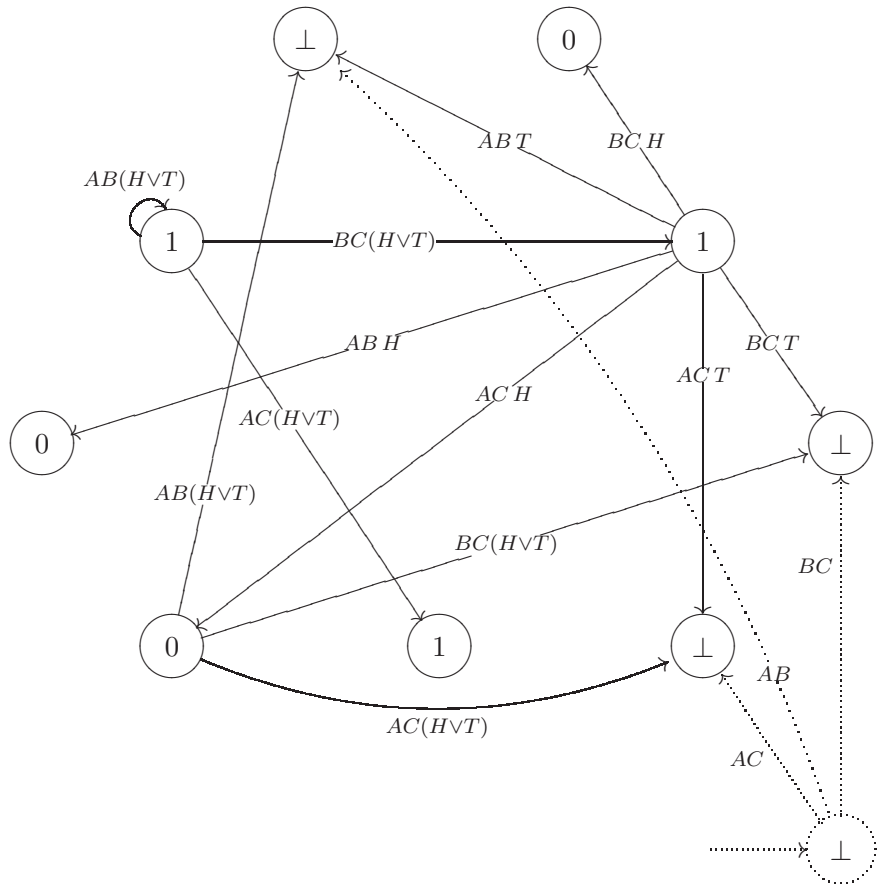

The overall structure of the Moore machine is as follows: The entry
states and transitions are dotted. Each of the three lines of three
states (they form roughly edges of a triangle), consists of states
with the same, already known pardon decision in the next experiment,
while the current experiment's outcome is represented as the label of
the state. Transitions are shown for one state on each edge only,
because their targets depend on the input only, and not on the source
within that edge.  And this is why we can calculate the probability of
\eqref{3pris} in a quite straightforward way. For time greater than
$1$ the probability of getting in two steps to a state with a given
label does not depend on the current state nor on the
time. Essentially, after the first step the edge of the triangle is
chosen, which corresponds to the move to one of the states in the
middle column of Fig.\ \ref{f7}. In the second step we move to the
state with the label equal to the destination label from Fig.\
\ref{f7}, and the edge it is found within depends on the next
experiment, already. The similarity is even stronger if we compare
Fig. \ref{f9} with Fig.\ \ref{3prisRR} rather than with Fig.\
\ref{f7}. A formal calculation, using matrix calculus, can be found in
Section \ref{algorytmy} below.

The most substantial difference is that \eqref{3pris} is not a
``Russian roulette'' model! To note this set time to $3$ and see: the
present outcomes depend on the pardon decisions made at time $2$,
while the Guard was testifying in the previous round of the
experiment, and while we are hearing the testimony of the Guard now,
the pardons are already decided as a part of the next experiment. So
the probabilistic choices which we described as irrelevant for the
Moore machine model, are parts of the previous/next repetition schema
here. The overlapping experiments do not interfere, however, so this
does not affect probabilities.  Furthermore, all the final outcome
undefined values have been merged into one state. Finally, there are
entry states which are visited just once and correspond to the
situation at time $1$, when the Guard says something, but there is no
pardon decision to compare it with.

A modified version of \eqref{3pris}, which is Russian roulette, is as
follows: 

\begin{equation}\label{3prisRR}
\left(
(\atone AB)
\lor
(\atone AC)
\left|
((\atone AB)\land \attwo(H\lor T))
\lor
((\atone BC)\land \attwo H)
\right.\right),
\end{equation}

where $\atone\alpha$ is $\PDIA(\lnot\PREV\TRUE\land \alpha)$ and
$\attwo\alpha$ is $\PDIA(\PREV\TRUE\land \lnot\PREV\PREV\TRUE\land
\alpha),$ and express that $\alpha$ is true at time $1$ and $2$,
respectively.

\begin{figure}[!hbtp]
\[\UseTips
\xymatrix @C=20mm
{
&&&*+++[o][F]{1}
\ar@(ur,dr)[]
\\
&&*+++[o][F]{{\bot}}
\ar[ur]^{\textstyle{H\lor T}}
&*+++[o][F]{{0}}
{ \ar@(ur,dr)[]}
\\
{\ar[r]}
&*+++[o][F]{{\bot}}
\ar@/^/[ur]|{\textstyle{AB}}
\ar[r]|{\textstyle{BC}}
\ar@/_/[dr]|{\textstyle{AC}}
&*+++[o][F]{{\bot}}
\ar[ur]|(.3){\textstyle{H}}
\ar[dr]|{\textstyle{T}}
\\
&&*+++[o][F]{{\bot}}
\ar@/_/[r]_{\textstyle{H\lor T}}
&*+++[o][F]{{\bot}}
\ar@(ur,dr)[]
}
\]
\caption[]{Moore machine of \eqref{3prisRR}. It is the minimalization
of the Moore machine from Fig.\ \ref{f7}, so they are indeed logically
indistinguishable.}
\label{f3prisRR}
\end{figure}
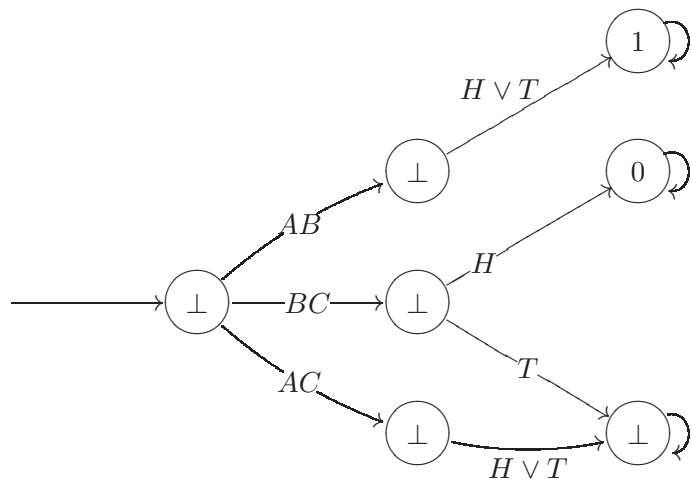

The general conclusion is that simple Moore machines can correspond to
complicated $\TT$ formulas, and simple $\TT$ descriptions can yield
complicated Moore machines. If we additionally take into account that
it is hard to expect that any computer program will be ever able to
transform human-readable representations of one kind into
human-readable representations of the other kind\footnote{In both
cases even graphical layout can have a huge impact on the readability
of the model!}, we recommend that the whole process of modelling is
done using only one of the formalisms, without mixing them.
 
\subsection{Algorithm for calculating the probability}\label{algorytmy}

Of course, the natural method to compute probability of a given
regular conditional $c$ in our model is to refer to an underlying
Markov chain $\X,$ perform the computations there, and then use the
formula

\[\Pr(c)=\frac{\sum_{i:h(i)=1}\lim_{n\to\infty}\Pr(X_n=i)}
{\sum_{i:h(i)=1~\text{or}~0}\lim_{n\to\infty}\Pr(X_n=i)},
\]

which follows directly from the Bayes' Formula.

The calculation of $\lim_{n\to\infty}\Pr(X_n=i)$ is generally known to
be polynomial time in the number of states of the Markov chain,
assuming unit cost of arithmetical operations \cite{KS}. The book
\cite{stewart} contains the account of state-of-the-art algorithms
for numerical calculations of the limiting probabilities.

As an example we calculate here the probability of the formula
\eqref{3pris}, using the simplest possible approach, assuming that all
the events from $\Omega$ have nonzero probability.

We assume the following numbering of the states of the Markov chain
from Fig.\ \ref{f9}:

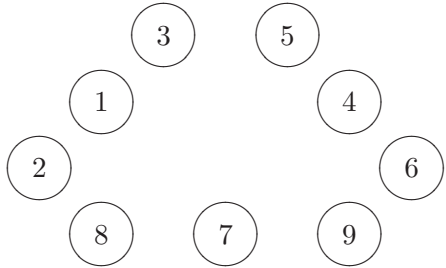
\begin{figure}[!hbtp]
\[
\xymatrix @R=0mm @C=0mm
{
&&*+++[o][F]{3}
&&*+++[o][F]{5}
\\
&*+++[o][F]{1}
&&&&*+++[o][F]{4}
\\
*+++[o][F]{2}
&&&&&&*+++[o][F]{6}
\\
&*+++[o][F]{8}
&&*+++[o][F]{7}
&&*+++[o][F]{9}
}
\]
\caption[]{Numbering of the states of Markov chain resulting from the
Moore machine in Fig.\ \ref{f9}.}\label{f10}
\end{figure}

Then the matrix $\Pi$ of transition probabilities is
\[
\begin{bmatrix} 
\bar{AB}& 0& 0& 
\bar{BC}&0 & 0&
\bar{AC}&0 & 0\\
\bar{AB}& 0& 0& 
\bar{BC}&0 & 0&
\bar{AC}&0 & 0\\
\bar{AB}& 0& 0& 
\bar{BC}&0 & 0&
\bar{AC}&0 & 0
 \\ 
0& \bar{AB}\,\bar{H}& \bar{AB}\,\bar{T}&
0& \bar{BC}\,\bar{H}& \bar{BC}\,\bar{T}&
0& \bar{AC}\,\bar{H}& \bar{AC}\,\bar{T}\\
0& \bar{AB}\,\bar{H}& \bar{AB}\,\bar{T}&
0& \bar{BC}\,\bar{H}& \bar{BC}\,\bar{T}&
0& \bar{AC}\,\bar{H}& \bar{AC}\,\bar{T}\\
0& \bar{AB}\,\bar{H}& \bar{AB}\,\bar{T}&
0& \bar{BC}\,\bar{H}& \bar{BC}\,\bar{T}&
0& \bar{AC}\,\bar{H}& \bar{AC}\,\bar{T}
 \\ 
0& 0& \bar{AB}&
0& 0& \bar{BC}&
0& 0& \bar{AC}\\
0& 0& \bar{AB}&
0& 0& \bar{BC}&
0& 0& \bar{AC}\\
0& 0& \bar{AB}&
0& 0& \bar{BC}&
0& 0& \bar{AC}
      \end{bmatrix}
\]
where $\bar{AB}$ stands for $\pr(AB),$ and similarly for arguments
$BC, AC, H,T$ (the matrix does not fit into the page when the standard
notation is used).

It can be directly checked that the square of this matrix has all
entries positive, hence the whole represents a single ergodic
class. (This is what breaks down when some elements from $\Omega$ have
probability $0.$ It this is permitted, one has to consider a few more
cases.) It is known that in such cases the limiting probability does
not depend on the initial probabilities of getting into this class,
therefore we can ignore the dotted (transient) states from
Fig.~\ref{f9}. The limiting probabilities can be found, given
$\Pi=(p_{ij}),$ by finding the only solution of the system of linear
equations

\[\left\{\begin{array}{lcl}
\sum_{i=1}^9x_i&=&1,\\[5pt]
\sum_{i=1}^9p_{i1}x_i&=&x_1,\\[5pt]
\sum_{i=1}^9p_{i2}x_i&=&x_2,\\[5pt]
\cdots&=&\cdots\\[5pt]
\sum_{i=1}^9p_{i9}x_i&=&x_9,
         \end{array}\right.
\]

which yields the following unique solution:

\begin{align*}
x_1&=\bar{AB}^{2}&
x_2&=\bar{BC}\,\bar{AB}\,\bar{H}&
x_3&= \bar{AB}(1-\bar{AB} - \bar{BC}\,\bar{H})\\
x_4&=\bar{AB}\,\bar{BC}&
x_5&=\bar{BC}^{2}\,\bar{H}&
x_6&= \bar{BC}(1-\bar{AB} -\bar{BC}\,\bar{H})\\
x_7&=\bar{AC}\,\bar{AB}&
x_8&=\bar{BC}\,\bar{AC}\,\bar{H}&
x_9&= 1 - \bar{AB}(1 + \bar{AC} + \bar{BC}\,\bar{H})- \bar{BC} 
\end{align*}

and the asymptotic probability of the conditional represented by the
Moore machine in question is $\dfrac{\Pr(AB)}{
\Pr(BC)\Pr(H)+\Pr(AB)},$ as expected. In particular, in the
equiprobable case the value is $2/3.$

\section{Related work and possible extensions}

\subsection{Related work}\label{related}

\begin{itemize}
\item
Using temporal logic in reasoning about knowledge is nothing
new. Indeed, many logics of knowledge incorporate temporal operators,
see \cite{Fagin}. However, to the best of our knowledge, $\TT$ is the
very first multi-valued temporal logic to be considered. In
particular, the above mentioned logics of knowledge are two-valued.
Moreover, $\TT$ is the first natural use of past tense temporal logic
in computer science. Most of the established formalisms which use
propositional temporal logic, indeed use its future tense fragment.

\item
Computing of conditional probabilities $\Pr(\varphi|\psi)$ is not new,
either, and has been considered by several authors, including
\cite{liogonkii,halp1,halp2}, mostly for first order logic of
unordered structures.

\item
Finally, Markov chains have already been used for evaluation of
probabilities of logical statements. In particular, our Bayes' Formula
is a simple extension of a theorem of Ehrenfeucht (see \cite{lynch}),
phrased there as a theorem about first order logic of ordered unary
structures (over which first order logic is equally as expressive as
propositional temporal logic, see \cite{Emerson}).
\end{itemize}

\subsection{Possible extensions.}

\begin{itemize}
\item
$\TT$ is not closed under its own connectives, since the nesting of
the conditioning operator $(\cdot|\cdot)$ with other connectives (let
alone itself) is not allowed, and since the temporal connectives
cannot be applied to a conditional pair. As a consequence, operations
on conditionals are defined by disassembling the pairs and
reassembling them afterwards, to yield a pair in the correct
syntactical form again.

We would like to have an equivalent logic with much better syntactical
structure. This should be possible by extending the ideas of
multivalued modal logics, investigated in \cite{mo,fit1,fit2}, by a
multivalued counterparts of $\Since.$ The logic would then assume the
form of a propositional logic with multivalued temporal connectives
and conditioning.

The big question is whether one can retain the Bayes' Formula
then. The existing attempts in the present tense logics of
conditionals suggest it might be difficult.

\item 
$\TT$ does not match exactly the class of automata, which for any
assignment of probabilities yield a Markov chain with all states
either transient or aperiodic. In such Markov chains all the limiting
probabilities do exist, and thus every such Markov chain can be
meaningfully considered to represent an extended kind of a
conditional. Indeed, below is a simple example of such an automaton.

\begin{figure}[!hbtp]
\[\UseTips
\xymatrix
{
*++[o][F]{~}
\save[]+<-8mm,-4mm>*{}\ar[]\restore
\ar@(l,u)^a
\ar@/^/[r]^{a^\C}
&*++[o][F]{~}
\ar@(u,r)^{a^\C}
\ar@/^/[d]^{a}
\\
*++[o][F]{~}
\ar@(d,l)^{a^\C}
\ar@/^/[u]^{a}
&*++[o][F]{~}
\ar@(r,d)^{a}
\ar@/^/[l]^{a^\C}
}
\]
\caption[]{It is not hard to verify
that, no matter what probability is assigned to the event $a,$ the
resulting Markov chain has only transient and acyclic states. However,
the automaton is not acyclic, since it has two states, reachable by a
path labelled $aa^\C$ from each other.}
\end{figure}
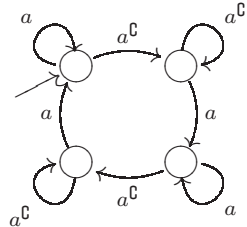

We would like to have an extension of $\TT,$ matching exactly the
class of Markov chains with only transient and aperiodic states, to
take the advantage of the maximal class of Markov chains for which the
limiting probabilities exist, and thus all the definitions given in
the paper make sense. We expect the logic to be obtained by extending
the multivalued temporal logic proposed suggested above, rather than
by extending the present syntax.

\end{itemize}

\paragraph{Acknowledgement.} The first author wishes to thank Igor
Walukiewicz for valuable informations concerning temporal logic.

\newpage

\bibliographystyle{abbrv} \bibliography{bib}


\end{document}